\documentclass[letterpaper, 10 pt, conference]{ieeeconf}  

\IEEEoverridecommandlockouts                              

\usepackage{amsmath}                
\usepackage{amssymb}                
\usepackage[dvipsnames]{xcolor}                 
\usepackage{bm}         
\usepackage{graphicx}
\usepackage{booktabs}

\usepackage{comment}
\usepackage{algpseudocode}
\usepackage{algorithm}
\usepackage{amsfonts}

\usepackage{tikz}

\usetikzlibrary{
    arrows.meta,
    calc
}

\usepackage{subcaption} 
\let\vec\bm

\usepackage{hyperref}
\usepackage[capitalise]{cleveref}

\makeatletter
\let\oldsine\sin
\let\oldcos\cos

\renewcommand{\sin}{\@ifnextchar\bgroup\sin@arg\oldsine}
\renewcommand{\cos}{\@ifnextchar\bgroup\cos@arg\oldcos}

\newcommand{\sin@arg}[1]{s_{#1}}
\newcommand{\cos@arg}[1]{c_{#1}}
\makeatother

\title{\LARGE \bf
Resilient Motion Planning for Free-Flying Space Robots \\under Actuator Failures }

\author{Nicolas de Maddalena$^{1,*}$, Joris Verhagen$^{2,*}$, Jana Tumova$^{2}$
\thanks{$^{*}$ Contributed equally to this work}
\thanks{$^{1}$ Institute for Dynamic Systems and Control, ETH Zürich, 8092 Zürich, ZH, Switzerland
        {\tt\small nicolade@student.ethz.ch}}%
\thanks{$^{2}$ Division of Robotics, Perception and Learning, KTH Royal Institute of Technology, Stockholm, Sweden, and also affiliated with Digital Futures.{\{\tt\small jorisv, tumova\}}
		{\tt\small @kth.se}}%
}

\begin{document}

\maketitle

\thispagestyle{empty}
\pagestyle{empty}


\begin{abstract}
Free-flying robots rely on multiple thrusters to maneuver in space. 
If one or more of these thrusters fail, the robot may lose control authority and risk mission failure. 
At the same time, their free-flying nature implies that, even in the absence of actuation, they continue along (locally) straight-line trajectories.
In this work we present a probabilistic, proactive, motion planning framework that explicitly accounts for actuator failures in space.
We model actuator failure modes as a Markov chain and propagate the probability of successfully reaching the goal along the planning horizon.
Precomputed reachable sets evaluate the robot's capabilities of reaching waypoints under potential failures and an RRT$^*$-based planner concatenates these waypoints.
The resulting algorithm maximizes the overall target-reaching probability, providing maximally resilient motion plans utilizing free-flying properties.
We validate our approach experimentally on a physical free-flyer platform with injected actuator failures. \href{https://github.com/nicolade119/RAF_RRTstar}{code and demos}
\end{abstract}

\section{INTRODUCTION}
\label{sec:intro}
Autonomous spacecraft operate in environments that combine high scientific and economic value with extreme inaccessibility. 
As such, failures to critical subsystems are typically irreversible once a mission is deployed.
Among such failures, actuator faults constitute a major source of mission degradation or complete mission loss, accounting for approximately 30\% of attitude and orbit control system failures~\cite{tafazoli2009study}.
As the number of autonomous spacecraft grows, ensuring safety and mission completion in the presence of actuator failures is a central challenge.

Traditionally, resilience to subsystem failures is achieved through electrical and mechanical redundancies such as reconfigurable components and duplicated actuators~\cite{post2021modularity} as well as improved testing~\cite{bouwmeester2022improving,hasan2022fault}. 
While effective, redundancy increases mass, complexity, and cost of a system that may still fail.
Existing fault-tolerant strategies are often reactive: upon detecting a failure, the system transitions to a safe mode or continues operation with reduced performance. 
Both approaches respond to failures after they occur, leading to greedy behavior that compromises long-horizon objectives.

The objective of this paper is instead to develop a proactive resilient motion planning algorithm in the presence of actuator failures, specifically for autonomous space robots.
Our approach exploits two key properties often found in space robotics: actuator symmetry and free-flying dynamics.
By reasoning over long planning horizons, we trade nominal performance for the ability to complete the mission after the loss of one, several, or all actuators.
While a spacecraft may not survive \emph{every} failure realization, we seek to maximize the probability of mission success over all possible failure realizations, providing a quantitative measure of the system's resilience w.r.t. the environment and task.

\setlength{\belowcaptionskip}{-10pt}
\begin{figure}
    \centering
    \includegraphics[width=1.0\linewidth]{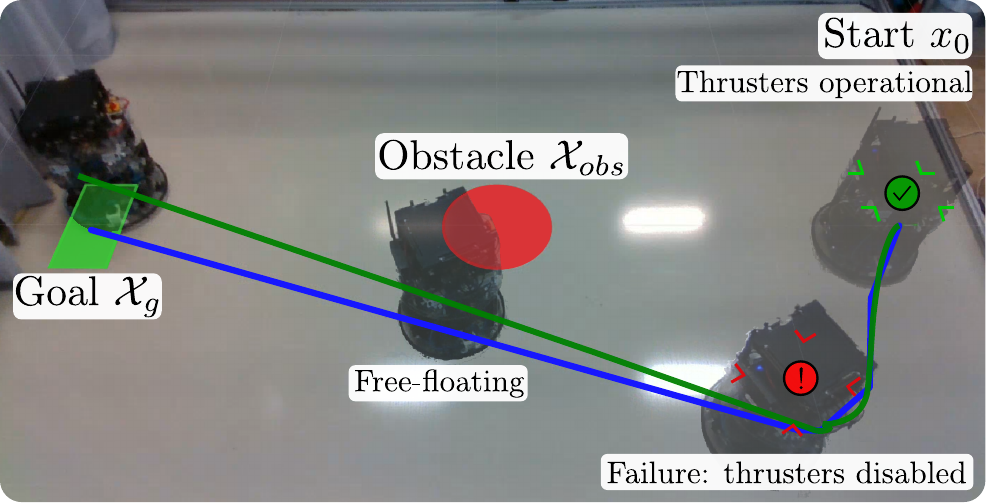}
    \caption{Planned (\textcolor{blue}{blue}) and executed (\textcolor{green}{green}) trajectory of a spacecraft analogue platform tasked with reaching the goal $\mathcal{X}_g$ while avoiding obstacle $\mathcal{X}_{obs}$. Complete failure occurs at the corner. The plan accounts for this possibility by escaping the obstacle's shadow, after which the remaining trajectory is a straight line to the goal that a total actuation loss can complete.}
    \label{fig:experiment_mid}
    \vspace{-3mm}
\end{figure}

We propose a motion planning framework integrating reachability analysis and Markov chains in an RRT$^*$ structure. 
Deterministic reachable sets are used to certify feasibility under specific actuator configurations, while a Markov chain models stochastic actuator failure events. 
Our approach explicitly incorporates the symmetry of actuator configurations found on spacecraft and the (local) free-flying dynamics that enable mission continuation even under complete loss of control authority.
Each sampled state is assigned a probabilistic reachability guarantee under the different failure scenarios, and entire trajectories inherit a reach-avoid probability for the task.
An example, shown in \cref{fig:experiment_mid}, shows how a plan requires the robot to leave the obstacle's shadow from the goal as early as possible and to keep its velocity vector aimed at the goal from that moment onward, enabling goal-reaching with full or partial thruster loss.


\subsection{Contributions}
\label{ssec:contributions}
We list our contributions as follows:

\begin{itemize}
    \item We present a motion planning framework for non-linear dynamical systems under uncertain actuator failures obtaining probabilistic reachability guarantees.
    \item Our planner maximizes resilience to actuator failures in the presence of symmetry and free-flying dynamics via a modified RRT$^*$ algorithm.
    \item We validate on the free-floating ATMOS platform~\cite{roque2025towards}.
\end{itemize}

\subsection{Related Work}
\label{ssec:related_work}
The field of resilient robotics has seen significant progress in model-based planning and fault-tolerant control. However, the literature often addresses system uncertainty and discrete structural failures separately.

Traditional approaches to robust planning often focus on stochasticity and parametric uncertainty. In \cite{nakka2022trajectory} planning for stochastic systems under uncertainty is tackled, yet this framework does not allow for discrete structural changes inherent to actuator failures. 
Similarly, domain randomization and robust optimization techniques~\cite{lin2022distributionally} mitigate the impact of disturbances and parametric shifts. However these methods provide empirical robustness rather than long-horizon formal probabilistic guarantees and are similarly less well-suited to handle actuator failures.

When failures do occur, reactive strategies are the standard. Research into fault-tolerant control~\cite{blanke2000fault,stockner2025fault,bayer2007planning,hasan2022fault} provide best-effort measures after a fault is detected. While these reactive methods are comparable to our work, they lack proactive foresight into choosing trajectories that minimize risk of mission failure after actuator failures occur.

In~\cite{vega2025contingency} safety guarantees for satellites under full actuator failures are addressed by staying in unstable regions of the orbit. While our work does not consider orbital dynamics, it does allow for independent failures of multiple thruster configurations.
Alternative approaches utilize learning-based, model-free~\cite{strombergermodel}, and model-based~\cite{liu2024reinforcement} setups to adapt to failures. While powerful, these tools lack the formal statistical guarantees required for high-stakes space missions.

Most close to our approach is \emph{path-tree optimization} in~\cite{phiquepal2022path} to balance performance with expected discrete outcomes, in this case observability resulting in belief-space graphs. We extend this philosophy to the domain of actuator failures, essentially creating probabilistic operability graphs.

\section{PRELIMINARIES}
Let $\mathbb{R}$ denote the set of real numbers, $\mathbb{B}=\{0,1\}$, and $[0,1]^n$ the set $\{\vec{x}\in\mathbb{R}^n \mid 0\leq x_i\leq 1,~\forall i\in\{1,\ldots,n\}\}$.
Vectors are denoted by bold lowercase letters, $\vec{x}$, matrices by bold capital letters, $\vec{X}$, and sets by calligraphic letters, $\mathcal{X}$.
A tilde distinguishes a node of the planning tree, $\tilde{x}$, from the state $\vec{x}$ that it carries.
We denote a timestep with index $k$ and a failure mode with index $i$.


\subsection{Markov Chains}
\label{ssec:markov_chain}
A Markov chain~\cite{markov1906rasprostranenie,howard2012dynamic} models stochastic transitions between discrete system states and is defined by the pair $(\mathcal{S},\vec{P})$, where $\mathcal{S}=\left\{s_0, s_1, \ldots, s_{m-1}\right\}$ is a finite set of states and $\vec{P}\in\left[0,1\right]^{m\times m}$ is a transition matrix with entries
\begin{equation}
    \label{eqn:markov_trans_prob_mat_entry}
    p_{a,b} = \mathbb{P}(s_{k+1}=s_a \mid s_k=s_b),
\end{equation}
i.e. the probability of moving to $s_a$ given that the chain occupies $s_b$ at timestep $k$.
Each column of $\vec{P}$ is therefore a probability distribution over $\mathcal{S}$, i.e. $\sum_{a=0}^{m-1} p_{a,b} = 1,~\forall b$.
Assigning to each timestep a probability vector $\vec{p}_k\in[0,1]^m$, whose $i$-th entry is the probability of occupying $s_i$, the chain propagates as $\vec{p}_{k+1} = \vec{P}\vec{p}_k$ and, by the Markov property, $\vec{p}_k = \vec{P}^k\vec{p}_0$.

\subsection{Hamilton-Jacobi Reachability}
\label{ssec:back_hj_reach}
Hamilton-Jacobi (HJ) reachability is a tool for formally verifying safety and liveness properties of dynamical systems. 
We consider the dynamical model of the spacecraft in its general nonlinear form
\begin{equation}
    \dot{\vec{x}} = f(\vec{x},\vec{u})
\end{equation}
where $\vec{x} \in \mathcal{X} \subset \mathbb{R}^{n_x}$ and $\vec{u} \in \mathcal{U} \subset \mathbb{R}^{n_u}$ indicate the state and control input to the system. We can then define the Backwards Reachable Set (BRS) as
\begin{equation}
\label{eqn:BRS}
\begin{aligned}
    \mathcal{G}(t;t_f, \mathcal{X}_f) = \{\vec{x}(t) \mid\  &\exists \vec{u}(\cdot) \in \mathcal{U}, \\
    &\dot{\vec{x}}=f(\vec{x},\vec{u}), \vec{x}(t_f)\in\mathcal{X}_f\},
\end{aligned}
\end{equation}
which denotes all the states at time $t$ from which a target $\mathcal{X}_f$ can be reached at time $t_f$.
For nonlinear systems, obtaining this BRS amounts to solving the well-known Hamilton-Jacobi-Bellman Partial Differential Equation (PDE)
\begin{equation}
\label{eq:HJB}
\begin{aligned}
    \frac{\partial}{\partial t}V(t,\vec{x}) + &\min_{\vec{u} \in \mathcal{U}} \nabla V(t,\vec{x})^\top f(\vec{x},\vec{u}) = 0, \\ &V(t_f,\vec{x}) = v(\vec{x}),
\end{aligned}
\end{equation}
where $v(\cdot)$ is any Lipschitz function whose zero sublevel set is the target, $v(\vec{x})\leq 0 \Leftrightarrow \vec{x}\in\mathcal{X}_f$. The sign of the value function then indicates whether $\mathcal{X}_f$ is reachable from $(t,\vec{x})$, so that
\begin{equation}
    \mathcal{G}(t;t_f,\mathcal{X}_f) = \{\vec{x} \mid V(t,\vec{x}) \leq 0\}.
\end{equation}
The PDE is nonlinear and infinite-dimensional, generally not permitting closed-form solutions.
The level-set method~\cite{1463302} provides an algorithmic approach to obtain this value function via dynamic programming.
The optimal control is then the minimizer of the value function~\cite{bansal2017hamilton} and drives the system from the BRS to the target set
\begin{equation}
\label{eq:HJB_opt_control}
    \vec{u}^*(t,\vec{x}) = \arg \min_{\vec{u}\in\mathcal{U}}\nabla V(t,\vec{x})f(\vec{x},\vec{u}).
\end{equation}
We refer to~\cite{bansal2017hamilton} and ~\cite{evans1984differential} for a comprehensive overview of reachability and level-set methods.

\subsection[RRT* Motion Planning]{RRT$^*$ Motion Planning}
A rapidly-exploring random tree (RRT)~\cite{lavalle1998rapidly} samples states in the state space and adds them to a tree $G=(V,E)$ defined by its vertices $V$, i.e. the samples, and the edges $E$ between pairs of vertices. In the limit such a tree covers the whole planning space, making the algorithm probabilistically complete, i.e. the probability of finding a solution converges to 1.
While RRTs ensure global state coverage, a vertex never changes its parent once connected, which confines the planner to suboptimal solutions.
RRT$^*$~\cite{karaman2011sampling} addresses this limitation and guarantees asymptotic optimality by rewiring the edges between the nodes in order to reduce the cost (e.g. the path length from the root). 
\section{PROBLEM STATEMENT}
The problem addressed in this work falls under a classic reach-avoid trajectory optimization problem with the added challenge of providing probabilistic guarantees under uncertain actuator failures.
First, we define the goal region $\mathcal{X}_g$ and its reach-event time
\begin{equation*}
    t_g = \inf\{t: t \geq 0, \vec{x}(t) \in \mathcal{X}_g\}.
\end{equation*}
Let $\vec{p}(t) \in \mathbb{R}^m$ then denote the probability of reaching $\vec{x}(t)$ under $m$ failure modes $M_i, i\in\{0, \ldots, m-1\}$ (each with associated probabilities of occurrence).
We then denote the resilient motion planning problem as

\begin{subequations}
\label{eqn:opt_problem}
    \begin{align}
        \label{eqn:cost_func}
        J^* &= \max_{\vec{x},\vec{u}} \mathrm{w}_\mathrm{p} \cdot \sum\vec{p}(t_g) - \mathrm{w}_\mathrm{d} \cdot d(\vec{x}(t_g)) \\
        \label{eqn:dynamics}
        s.t.~&\dot{\vec{x}} = f(\vec{x}(t), \vec{u}(t)) \\
        & \vec{x}(t) \in \mathcal{X}_{free}, \hspace{1.5mm} \mathcal{X}_{free} = \mathcal{X} \setminus \mathcal{X}_{obs} \\
        & \vec{x}(t_g) \in \mathcal{X}_g \\
        & \vec{x}(0) = x_0 \\
        & \vec{u}(t) \in \mathcal{U}_{M(t)} \label{eqn:control_limits}
    \end{align}
\end{subequations}
where $f(\cdot)$ denotes the state dynamics, and $\mathcal{X}_{free}$ denotes the obstacle-free part of the state space $\mathcal{X}$. 
The failure modes are captured in the \emph{failure-model}-dependent actuator constraints $\vec{u}(t) \in \mathcal{U}_{M_i}$ which, combined with the continuous dynamics, leads to a hybrid nonlinear dynamical system with probabilistic transition. It is the state of $\mathcal{U}_{M_i}$ along the trajectory $\vec{x}(t)$ that determines $\vec{p}(t_g)$.


The function $d(\vec{x}(t_g))$ computes the traveled distance from the root $\vec{x}(0)$ to the target node $\vec{x}(t_g)$ along its trajectory $\vec{x}(t)$.
The weights $\mathrm{w}_\mathrm{p} > 0$ and $\mathrm{w}_\mathrm{d} > 0$ allow user-defined focus on resilience and (a metric for) efficiency.
The relative likelihood of the failure modes, $\vec{P}$, and the weights implicitly determine which of the modes the trajectory is shaped around.

\section{Method}
\label{sec:method}
In this section we present our sampling-based resilient motion planning algorithm to solve the continuous problem in \cref{eqn:opt_problem}. 
We embed the standard RRT$^{*}$ algorithm with the information on actuator failure probability and the reachability for those models for newly sampled points. This sampling allows us to consider the nonlinear hybrid dynamics in a global, tractable motion planning algorithm.


\begin{figure}
    \centering
    \includegraphics[width=\linewidth]{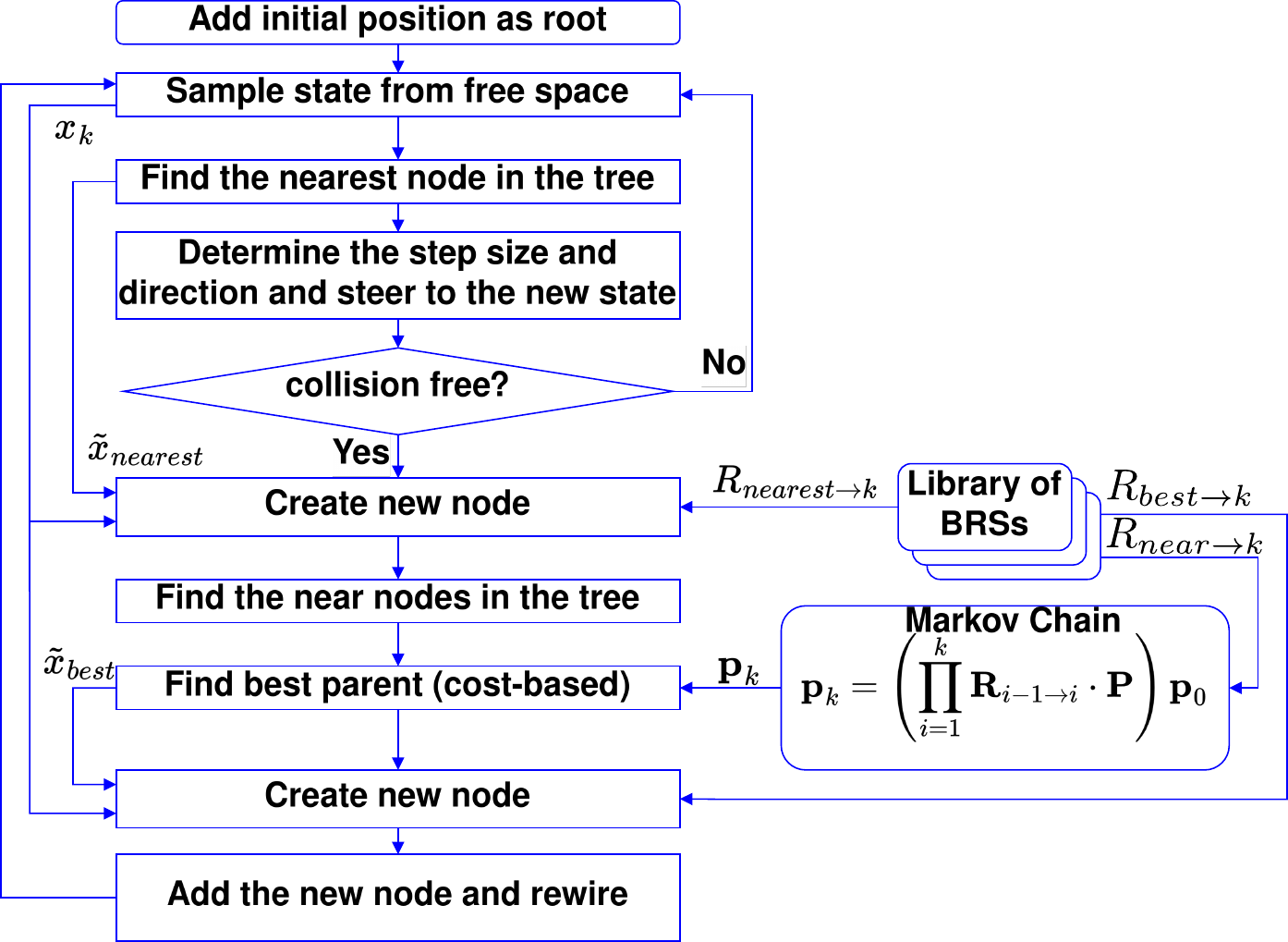}
    \caption{Flowchart of the resilient motion planning framework.}
    \label{fig:rrt*_flowchart}
\end{figure}

\subsection{Backwards Reachable Sets}
\label{ssec:method_brs}
First, according to \cref{eqn:dynamics} and \cref{eqn:control_limits}, we need to determine dynamic feasibility of a trajectory in the presence of potential actuator failures. 
We tackle this by considering the reachability between parent node $\tilde{x}_k$ and child node $\tilde{x}_{k+1}$ via the containment of $\vec{x}_k$ in the BRS of $\tilde{x}_{k+1}$.
Each of the $m$ failure models, captured by $\mathcal{U}_{M_i}, i\in\{0,\ldots,m-1\}$ leads to a different BRS.
Using the BRS, we guarantee the continuous-time satisfaction of \cref{eqn:dynamics} under a discrete occurrence of failure mode $M_i$ according to \cref{eqn:control_limits}. 
While this would lead to the assumption that actuator failures may only occur at the discrete samples, the consideration of subsystems ($\mathcal{U}_{M_i} \subseteq \mathcal{U}_{M_j}$) makes this less restrictive: any failure to a subsystem between $k$ and $k+1$ can be regarded as a failure of the subsystem at $k$.

Many spacecraft models exhibit some form of symmetry. 
The space-analogue platform in \cref{fig:atmos_fail_model} is actuated by eight symmetrically arranged thrusters and thus $\mathcal{U}_{M_2}$ is exactly a $90^\circ$ rotation of $\mathcal{U}_{M_1}$ (same $u_\theta$, swapped axes).
We can utilize this symmetry in two ways;
dynamically equivalent symmetric configurations reduce the number of BRSs that need to be computed and it reduces the number of connectivity/intersection checks in the algorithm.

\subsection{Failure Modeling}
\label{ssec:failure_modeling}
Let $\mathcal{M}=\{M_0, M_1, \ldots, M_{m-1}\}$ be the set of system models.
We assume that the failure probabilities (e.g. chance of failure per operating duration) are given. From that, a Markov chain can be constructed.
This chain's transition matrix $\vec{P} \in [0,1]^{m \times m}$ collects these probabilities.
To assess the probability of a trajectory $\vec{x}(t)$ reaching the goal under failure models, we extend the Markov chain with reachability information.
Let $\vec{R}_{k\rightarrow k+1} \in \mathbb{B}^{m\times m}$ be a diagonal matrix where the diagonal entry $\vec{R}_{k\rightarrow k+1}^{i} \in \mathbb{B}$ denotes whether $x_{k}$ is in the BRS of $x_{k+1}$ for model $i$.
We then denote the probability vector $\vec{p}_{k+1} \in [0,1]^m$ as the probability of $x_{k+1}$ being reachable by each failure model as $\vec{p}_{k+1} = \vec{R}_{k\rightarrow k+1}\vec{P} \vec{p}_k$ with $\vec{p}_k \in [0,1]^m$ the probability for being each failure model at $k$. For an entire trajectory consisting of sequential waypoints $(x_0,\cdots,x_k)$ we can obtain the probability vector
\begin{equation}
\label{eqn:failure_prop_markov}
    \vec{p}_{k} = \left(\prod_{j=k}^{1} \vec{R}_{j-1\rightarrow j}\cdot\vec{P}\right) \vec{p}_0 = \begin{bmatrix}
        p_k^0 & p_k^1 & \ldots & p_k^{m-1}
    \end{bmatrix}^\top
\end{equation}
recursively from the known root node probability $\vec{p}_0$, where element $p_{k}^{i}$ now denotes the probability of reaching node $x_k$ through model $M_i$.
Note that since $\vec{R}$ zeroes out non-traversible modes, $\sum_{i=0}^{m-1}p_k^{i} \leq 1$ is a sub-distribution. 
The deficit $1-\sum_{i=0}^{m-1}p_k^{i}$ is the probability of not reaching $x_k$ at all, i.e. the probability of failure.

\subsection{Sampling-based Motion Planner}
\label{ssec:methods_sbmp_rrt*}


To integrate the reachability and probability propagation into a unified framework, we employ a modified RRT$^*$ algorithm for motion planning under actuator failure uncertainty. 
The proposed planner incorporates the Markov failure probability matrix and precomputed backward reachable sets directly into its cost evaluation (see \cref{fig:rrt*_flowchart}). 
Throughout the remainder of this section, we focus on a planar system to illustrate the approach.

First, note that a sampling-based approach is well-suited to this problem due to its ability to handle high dimensional state spaces, non-convex environments, nonlinear dynamics, and discrete failure events. 
In contrast, trajectory optimization methods are often sensitive to initialization and prone to converge to local minima, while graph-based approaches suffer from the curse of dimensionality and require discretization incompatible with onboard computational constraints. 
Sampling-based algorithms such as RRT$^*$ are both probabilistically complete and asymptotically optimal~\cite{karaman2011sampling}. 
These guarantees formally hold only for monotonic cost functions. Our cost function needs to be slightly modified in order to be monotonic, because a local improvement for one trajectory candidate can still be worse than the trajectory candidate with the global minimal cost. 
To avoid converging to a local minimum, the algorithm stores the best cost found so far and ensures the final solution is no worse than any previous iteration.

The resilient RRT$^*$ planner is shown in \cref{alg:rrt*_failure}. 
First, nodes are created via $\textsc{CreateNode}(\cdot)$ with the properties of the current state, its parent, and the reachability matrix between parent and node. The root node does not have a parent or a reachability matrix. 
The node gets added to the tree (line~\ref{alg_line:init_V}) and all BRSs are loaded into memory for all models $M_i \in \mathcal{M}$ (line~\ref{alg_line:init_brs}).
We iteratively sample $N$ waypoints, $x_s \in \mathbb{R}^{n_x}$ uniformly from $\mathcal{X}_{\mathrm{free}}$ (line~\ref{alg_line:sample}).
We then continue to search for the nearest neighbor (line~\ref{alg_line:nearest}) in the reduced $(x,y,\theta)$ space using a KD-tree \cite{maneewongvatana1999analysisapproximatenearestneighbor}, significantly reducing the query time.

\begin{figure}
    \centering
    \includegraphics[width=\linewidth]{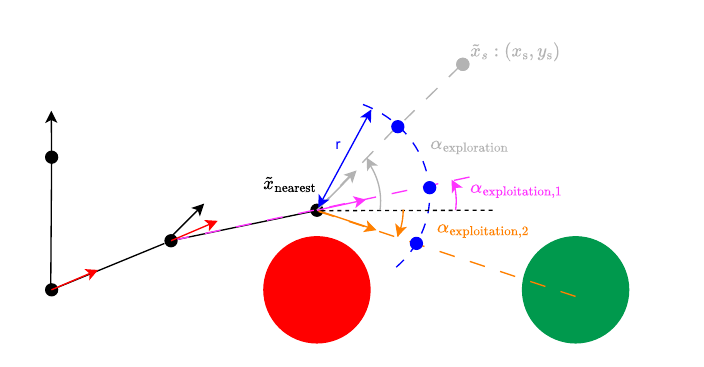}
    \caption{Angles and step size visualized for exploration and exploitation. The red area represents an obstacle. The green area is the goal region $\mathcal{X}_\mathrm{goal}$.
    The red arrows are the translational velocity vectors at the nodes in the tree. The black arrows represent the orientation at the node. The gray dot is the sampled point from $\textsc{SampleFree}(\cdot)$ and the blue points are the points resulting from the $\textsc{Steer}(\cdot)$ method. They are located at a random distance r (step size) from the nearest neighbor.}
    \label{fig:rrt*_explo}
\end{figure}

The function $\textsc{Steer}(\cdot)$ connects $\tilde{x}_\mathrm{nearest}$ to $x_s$, where we start to deviate from vanilla RRT$^*$~\cite{karaman2011sampling} (line~\ref{alg_line:steer}). The resulting node state $\vec{x}=(x,y,v_x,v_y,\theta)$ is sampled (exploration) or computed by heuristics (exploitation), where the steering direction $\alpha$ is chosen as
\begin{align*}
    \alpha_\mathrm{exploration} &= \operatorname{atan2}\left(y_s - y_n,x_s - x_n\right) \\
    \alpha_\mathrm{exploitation,1} &= \theta_n + \operatorname{atan2}\left(v_{y,n}^\mathcal{B},v_{x,n}^\mathcal{B}\right) \\
    \alpha_\mathrm{exploitation,2} &= \operatorname{atan2}\left(y_\mathrm{cg} - y_n,x_\mathrm{cg} - x_n\right),
\end{align*}
where the subscripts $s$ and $n$ denote the uniformly sampled state or the state of the nearest node respectively, with the state $\vec{x}=\begin{bmatrix} x & y & v_x & v_y & \theta \end{bmatrix}^\intercal$, and $(x_\mathrm{cg},y_\mathrm{cg})$ is the goal centroid. For a depiction of the different angles see \cref{fig:rrt*_explo}.
In short, $\alpha_{\mathrm{exploitation,1}}$ exploits free-flying capabilities of the spacecraft while $\alpha_{\mathrm{exploitation,2}}$ biases towards the goal.
The step size is uniformly sampled within the bounds derived from the BRS geometry.

Next, we check that the new connection is collision-free (line~\ref{alg_line:collision_free}) and if it is, we create the node data structure $\tilde{x}_k=\{\vec{x_k}, \tilde{x}_\mathrm{parent}, R_ \mathrm{parent \rightarrow k}\}$ (line~\ref{alg_line:create_node_1}). Similarly as in vanilla RRT$^*$, we search for near nodes (line~\ref{alg_line:near}). As mentioned above, we search only in the reduced $(x,y,\theta)$ space with a KD-tree structure. In order to improve the convergence rate we use the same shrinking sphere as~\cite{karaman2011sampling} with radius $r = \min\{\eta,\rho\}$ with
\begin{equation}
    \label{eqn:shrinking_radius}
    \rho = \gamma_{\text{RRT}^*}(\log(\text{card}(V))/\text{card}(V))^{1/d}
\end{equation}
where $\gamma_{\text{RRT}^*}$ can be computed as shown in~\cite{karaman2011sampling}.

We check if any of the near nodes $\tilde{X}_\mathrm{near}$ (line~\ref{alg_line:best_parent}) can be used as a parent for the new node $\tilde{x}_k$ by reducing its cost and update the node data structure (line~\ref{alg_line:create_node_2}).
Finally, we rewire~\cite{karaman2011sampling} the tree to reduce the cost locally (line~\ref{alg_line:rewire}).

\begin{algorithm}
\caption{RRT$^*$ - Failure}
    \begin{algorithmic}[1]
        \State \textbf{Initialize:} $\vec{x}_0, \mathcal{X}_g, \mathcal{X}_\mathrm{obs}$  \label{alg_line:init_env}
        \State $\tilde{x}_0 \gets \textsc{CreateNode}(x_0, \text{None}, \text{None})$  \label{alg_line:init_root}
        \State V $\gets \tilde{x}_0$  \label{alg_line:init_V}
        \State Load $\mathrm{BRS}^\mathrm{reach}_\mathrm{M_i}$  \label{alg_line:init_brs}
        
        \For{$k = 1, 2, \ldots, N$}{%
            \State $\vec{x}_s \gets \textsc{SampleFree}()$  \label{alg_line:sample}
            \State $\tilde{x}_\mathrm{nearest} \gets \textsc{NearestNode}(G=(V,E), \vec{x}_s$)  \label{alg_line:nearest}
            \State $\vec{x}_k \gets \textsc{Steer}(\tilde{x}_\mathrm{nearest}, \vec{x}_s, \mathrm{BRS}^\mathrm{reach}_\mathrm{M_0})$    \label{alg_line:steer}
            
            \If{$\textsc{CollisionFree}(\vec{x}_k, \tilde{x}_\mathrm{nearest})$}    \label{alg_line:collision_free}
                \State $\tilde{x}_k \gets \textsc{CreateNode}(\vec{x}_k, \tilde{x}_\mathrm{nearest}, R_\mathrm{nearest \rightarrow k})$  \label{alg_line:create_node_1}
                \State $\tilde{X}_\mathrm{near}$ $\gets$ $\textsc{NearNodes}$($G$, $\vec{x}_k$, $\min\{\eta,$ $\rho\})$    \label{alg_line:near}
                \State $\tilde{x}_\mathrm{best} \gets \textsc{FindBestParent}(\tilde{x}_\mathrm{k}, \tilde{X}_\mathrm{near})$    \label{alg_line:best_parent}
                \State $\tilde{x}_\mathrm{k} \gets \textsc{CreateNode}(\vec{x}_k, \tilde{x}_\mathrm{best}, R_\mathrm{best \rightarrow k})$  \label{alg_line:create_node_2}
                \State $V \gets V \cup \tilde{x}_k, \hspace{5pt} E \gets E \cup (\tilde{x}_\mathrm{best},\tilde{x}_k)$
                \State $\textsc{RewireTree}(\tilde{x}_\mathrm{k}, \tilde{X}_\mathrm{near})$    \label{alg_line:rewire}
            \EndIf
        }\EndFor
        \State \Return $G = (V, E)$
    \end{algorithmic}
    \label{alg:rrt*_failure}
\end{algorithm}

\subsection{Complexity}
\cref{alg:rrt*_failure} scales linearly with the number of models $m$ in both space and time complexity as each iteration $m$ set checks to fill the reachability matrix $\vec{R}_{k \rightarrow k+1}$. 
Since vanilla RRT$^*$ has $O(N\cdot \log(N))$ time complexity and $O(N)$ space complexity, our algorithm scales with $O(m\cdot N\cdot \log(N))$ and $O(m\cdot N)$, respectively.
Asymptotic optimality is maintained under a monotonic cost function which can be accomplished by defining the overall cost as the minimal-cost path in the tree as mentioned in Sec.\ref{ssec:methods_sbmp_rrt*}.
\section{Validation}
\label{sec:experiments}

We first show the performance of the planner in \cref{alg:rrt*_failure}. Next, we test the planner on a physical experimental space robot with injected actuator failures.

\begin{figure}
    \centering
    \begin{subfigure}[t]{0.37\linewidth}
        \centering
        \includegraphics[width=\linewidth]{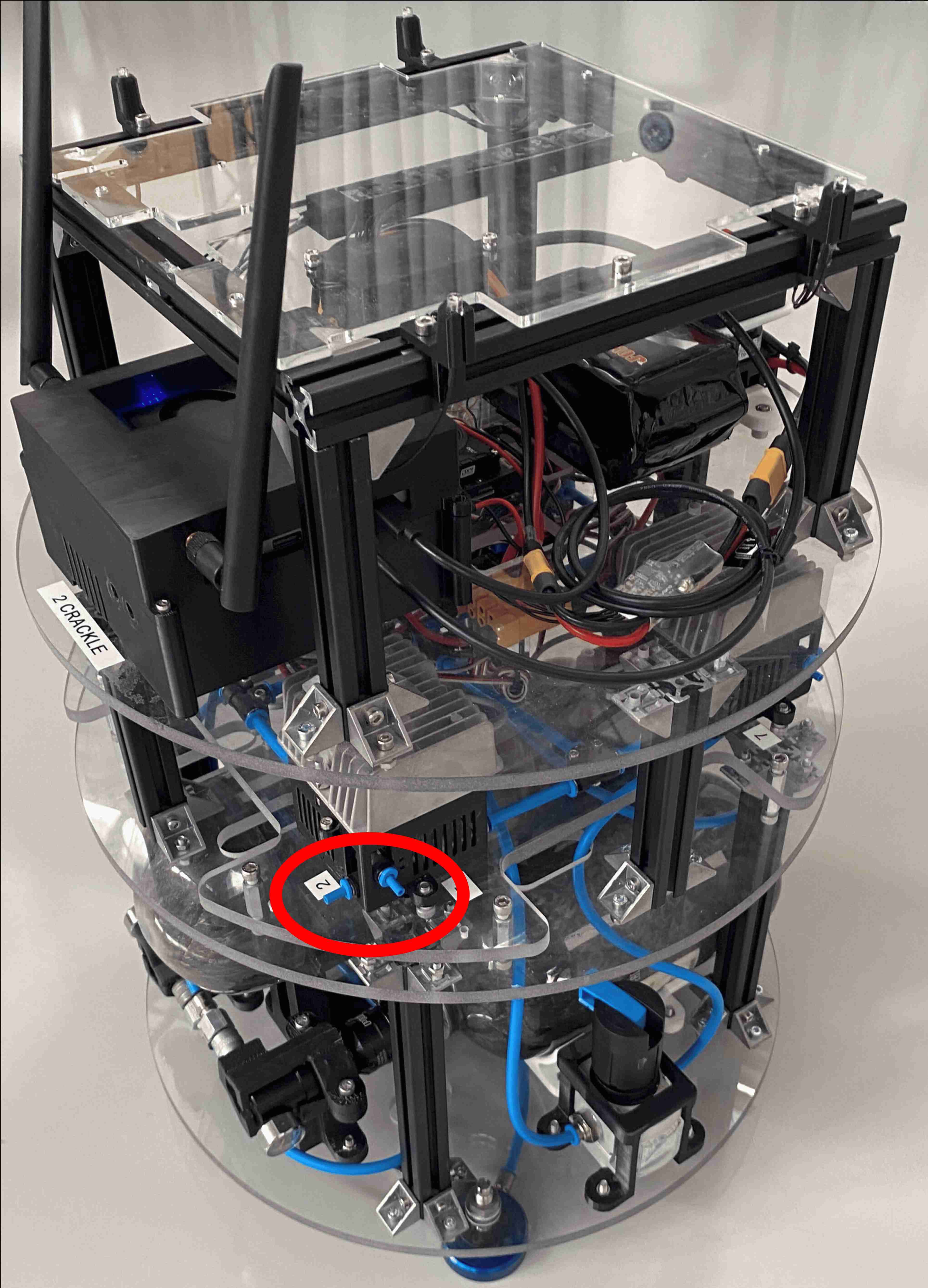}
        \label{fig:atmos}
    \end{subfigure}\hfill
    \begin{subfigure}[t]{0.62\linewidth}
        \centering
        \resizebox{\linewidth}{!}{%
        \begingroup

\tikzset{
    mode/.style={
        circle,
        draw=black,
        line width=1pt,
        minimum size=2.65cm,
        inner sep=0pt
    },
    transition/.style={
        draw=red,
        line width=1.1pt,
        -{Latex[length=3mm,width=2.2mm]}
    },
    transition label/.style={
        text=red,
        font=\Large,
        fill=white,
        inner sep=2pt
    },
    self loop/.style={
        draw=black,
        line width=1pt,
        -{Latex[length=3mm,width=2.2mm]}
    },
    self loop label/.style={
        font=\Large,
        fill=white,
        inner sep=1.5pt
    },
    robot body/.style={
        draw=black,
        line width=1pt
    },
    actuator/.style={
        draw=blue!85!black,
        line width=1.4pt
    }
}

\newcommand{\horizontalactuators}{%
    \draw[actuator] (-0.92, 0.54) -- (-0.54, 0.54);
    \draw[actuator] ( 0.54, 0.54) -- ( 0.92, 0.54);
    \draw[actuator] (-0.92,-0.54) -- (-0.54,-0.54);
    \draw[actuator] ( 0.54,-0.54) -- ( 0.92,-0.54);
}

\newcommand{\verticalactuators}{%
    \draw[actuator] (-0.54, 0.54) -- (-0.54, 0.92);
    \draw[actuator] ( 0.54, 0.54) -- ( 0.54, 0.92);
    \draw[actuator] (-0.54,-0.92) -- (-0.54,-0.54);
    \draw[actuator] ( 0.54,-0.92) -- ( 0.54,-0.54);
}

\newcommand{\drawrobot}[3]{%
    \begin{scope}[shift={(#1.center)}]

        \draw[robot body]
            (-0.73,-0.73) rectangle (0.73,0.73);

        \draw[robot body]
            (-0.35,-0.73) --
            ( 0.73, 0.00) --
            (-0.35, 0.73) -- cycle;

        \node[
            font=\LARGE,
            fill=white,
            inner sep=1.5pt
        ] at (-0.07,0) {#2};

        #3
    \end{scope}
}

\begin{tikzpicture}

    \node[mode] (M0) at ( 0.0, 3.25) {};
    \node[mode] (M1) at (-3.55,0.00) {};
    \node[mode] (M2) at ( 3.55,0.00) {};
    \node[mode] (M3) at ( 0.0,-3.25) {};


    \draw[transition]
        (M0) --
        node[
            transition label,
            midway,
            above left=2pt
        ] {$p_{0,1}$}
        (M1);

    \draw[transition]
        (M0) --
        node[
            transition label,
            midway,
            above right=2pt
        ] {$p_{0,2}$}
        (M2);

    \draw[transition]
        (M0) --
        node[
            transition label,
            midway,
            left=5pt
        ] {$p_{0,3}$}
        (M3);

    \draw[transition]
        (M1) --
        node[
            transition label,
            midway,
            below left=1pt
        ] {$p_{1,3}$}
        (M3);

    \draw[transition]
        (M2) --
        node[
            transition label,
            midway,
            below right=1pt
        ] {$p_{2,3}$}
        (M3);

    %

    \path[self loop]
        (M0) edge[
            loop left,
            min distance=17mm,
            looseness=5
        ]
        node[
            self loop label,
            left=2pt
        ] {$p_{0,0}$}
        (M0);

    \path[self loop]
        (M1) edge[
            loop above,
            min distance=17mm,
            looseness=5
        ]
        node[
            self loop label,
            above=2pt
        ] {$p_{1,1}$}
        (M1);

    \path[self loop]
        (M2) edge[
            loop above,
            min distance=17mm,
            looseness=5
        ]
        node[
            self loop label,
            above=2pt
        ] {$p_{2,2}$}
        (M2);

    \path[self loop]
        (M3) edge[
            loop left,
            min distance=17mm,
            looseness=5
        ]
        node[
            self loop label,
            left=2pt
        ] {$1$}
        (M3);


    \drawrobot{M0}{$M_0$}{%
        \horizontalactuators
        \verticalactuators
    }

    \drawrobot{M1}{$M_1$}{%
        \horizontalactuators
    }

    \drawrobot{M2}{$M_2$}{%
        \verticalactuators
    }

    \drawrobot{M3}{$M_3$}{}

\end{tikzpicture}

\endgroup%
        }
        \label{fig:failure_model}
    \end{subfigure}
    \vspace{-3mm}
    \caption{The space-analogue platform with its actuators and how we model the failure cases. The working actuators are depicted in blue and the failed actuators are erased from the drawing.}
    \label{fig:atmos_fail_model}
\end{figure}

The system under consideration is the experimental freeflyer platform in~\cref{fig:atmos_fail_model}.
This planar system is holonomic and has 8 actuators to move in 3 degrees of freedom. 
However, actuation failures may significantly alter the dynamic capabilities of the system, potentially making it underactuated, non-holonomic, or even uncontrollable.
We model four different failure cases, and therefore our transition probability matrix is $\vec{P}\in[0,1]^{4\times 4}$ with the probability vector $\vec{p}\in[0,1]^4$.
Model 0 ($M_0$) is the nominal model, i.e. all actuators are working and the system is over-actuated. Model 1 ($M_1$) and model 2 ($M_2$) are models in which four actuators fail simultaneously and are classified as a single failure event. For each, either the lateral actuators ($M_1$) or the longitudinal actuators ($M_2$) have failed. This limits the input and the robot is no longer fully actuated.
The last model, model 3 ($M_3$), is a double failure and has no working actuators left. It is a combination of the failures of $M_1$ and $M_2$. Refer to \cref{fig:atmos_fail_model} for a visual representation.

We model our actuators such that a failure corresponds to a complete loss of functionality. If failed, it becomes disabled, cannot exert any force onto the system, and remains inoperative for the mission duration.
Operational actuators are assumed to function without degradation or limitation.

While our method supports a higher number of failure models, it should be noted that considering each unique failure mode leads to $2^8=256$ possible combinations.
We instead opt for a set of failure models containing all other failure models: $M_1$ and $M_2$ capture all models up to 4 failures in the lateral or longitudinal direction while $M_3$ captures all models with multiple failures in any direction.

\subsection{Planner}


We showcase the planning algorithm in a single scenario, where we stop the algorithm after 2000 samples.
The total failure probability is assumed to be 1\%.
Furthermore, we assume a single failure is more likely to happen than a double failure. The transition probability matrix $\vec{P}$ used for the generation of the following trajectories is

\begin{equation*}
    \label{eqn:trans_prob_mat_eval}
    \vec{P} = \begin{bmatrix}
        0.99    &  0    & 0         & 0 \\
        0.00495 & 0.99  & 0         & 0 \\
        0.00495 & 0     & 0.99      & 0 \\
        0.0001  & 0.01  & 0.01      & 1 \\
        \end{bmatrix}
\end{equation*}

The target sets for the BRS computation in \cref{eqn:BRS} have different configurations which allow for faster convergence of \cref{alg:rrt*_failure}. Specifically, a target can be specified as a resting state, i.e. $\vec{x}_\mathrm{target,static}=\begin{bmatrix} x & y & 0 & 0 & \theta \end{bmatrix}$, or a moving state, i.e. $\vec{x}_\mathrm{target,moving}=\begin{bmatrix} x & y & v_x & v_y & \theta \end{bmatrix}$. To reduce the number of possible target states, we specify a single moving target velocity $\overline{v}_\mathrm{target,moving}=\sqrt{v_x^2 + v_y^2} = 0.15\frac{m}{s}$. 
Additionally, we specify that the system must be in a configuration where $v_x = v_y$ (which are in the body frame), meaning that in case of a $M_1$ or $M_2$ failure the system has equal control authority to slow down or accelerate.
We compute the backwards reachable sets using OptimizedDP~\cite{https://doi.org/10.48550/arxiv.2204.05520} which obtains state-of-the-art performance and accuracy.

Specifying the goal as a resting state would prevent $M_3$ (the platform with all actuators failed) from ever reaching this goal. It can however be advantageous in certain space applications, for instance when the spacecraft needs to receive or send a signal that is only possible in that region.
To that end, we allow goal-reaching purely on position for $M_3$.

\begin{figure}
    \centering
    \begin{subfigure}[t]{0.5\linewidth}
        \centering
        \includegraphics[width=\linewidth]{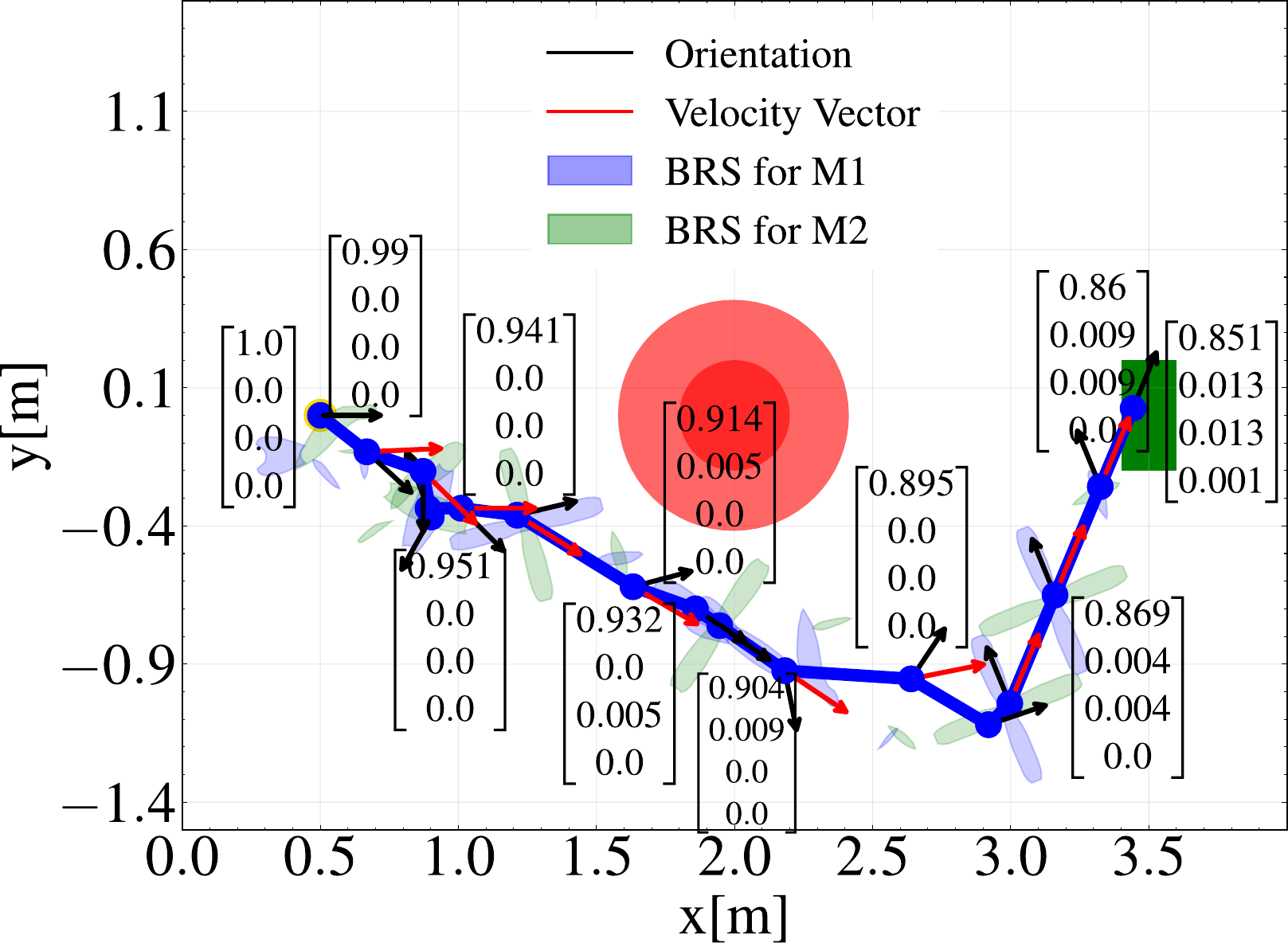}
    \end{subfigure}\hfill
    \begin{subfigure}[t]{0.5\linewidth}
        \centering
        \includegraphics[width=\linewidth]{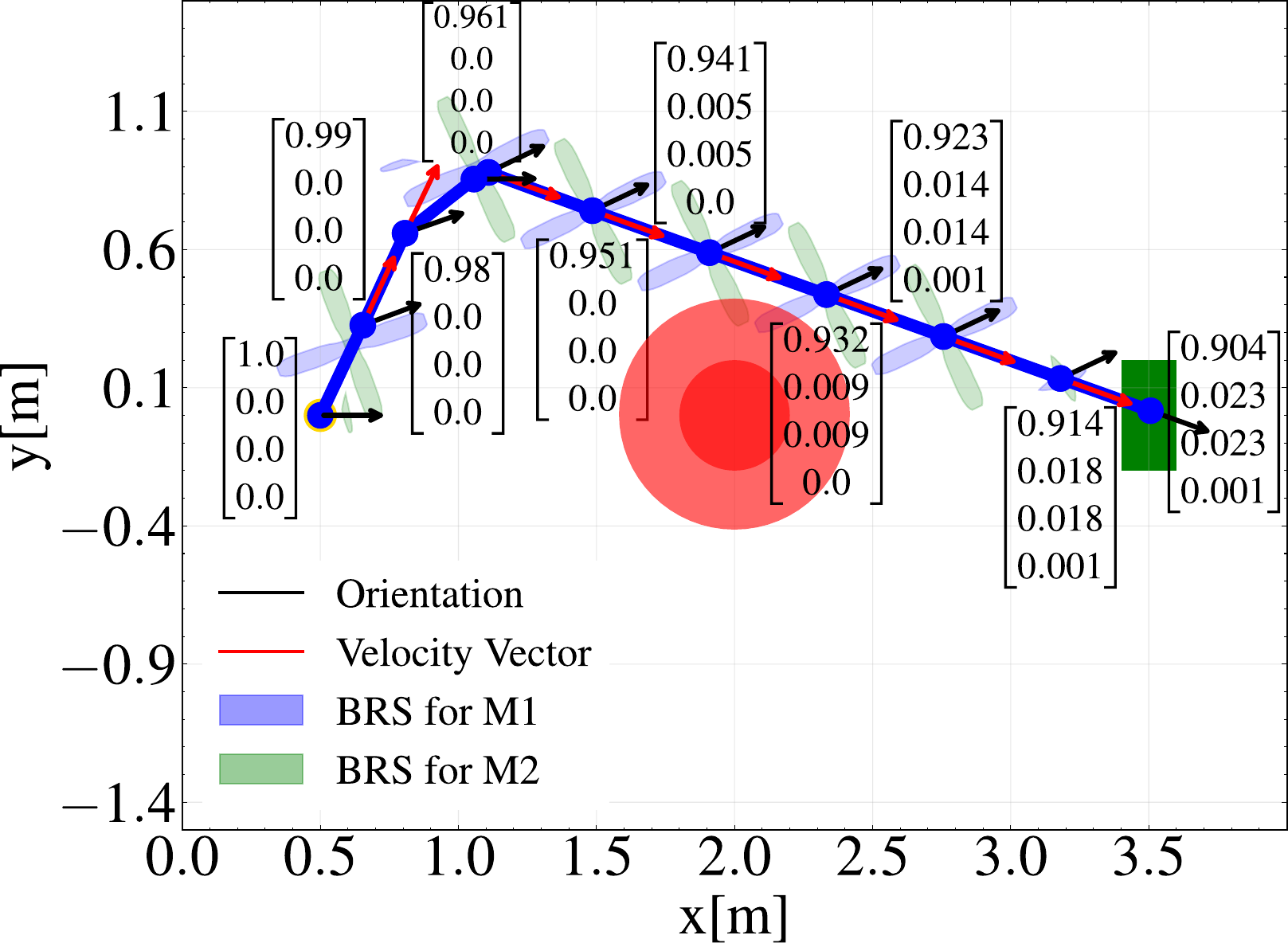}
    \end{subfigure}
    
    \caption{Intermediate (Left) and Final (Right) solution of a motion plan around an obstacle (denoted in red) towards a goal region (denoted in green) using~\cref{alg:rrt*_failure}. The state of a node is represented by a dot for position $(x,y)$, a black arrow for orientation $(\theta)$ and a red arrow for velocity $(\vec{v})$. The red shadows around the obstacles account for the size of the spacecraft and the avoidance BRSs of $M_1$ and $M_2$. 
    The BRS for $M_1$ (lateral failure) and $M_2$ (longitudinal failure) are displayed in blue and green. The BRS for $M_0$ (nominal) and $M_3$ (all actuators failed) are not shown for clarity. 
    }
    \label{fig:mid_traj}
\end{figure}

The trajectory in \cref{fig:mid_traj} shows how the probability of reaching the goal is improved iteratively.
Due to the straight-line steering in line~\ref{alg_line:steer} of \cref{alg:rrt*_failure}, the planner accounts for the free-flying characteristic of $M_3$ and obtains a trajectory which leads to the goal area in a straight-line.
Note that therefore $\sum\vec{p}_k=\sum\vec{p}_{k+1}$ if $\vec{x}_{k+1}$ lies on the straight-line constant-velocity extension of $\vec{x}_{k}$.

The final solution has $\vec{p}(t_g)=[0.904,\,0.023,\,0.023,\,0.001]^\intercal$, i.e. a success probability of $\sum\vec{p}(t_g)=0.951$, against $0.878$ for the intermediate solution.
The final plan loses probability only during the initial maneuver ($1.0\rightarrow0.99\rightarrow0.98\rightarrow0.961\rightarrow0.951$) and preserves $0.951$ over every subsequent edge to the goal.
This is the trade-off of \cref{eqn:cost_func}: the planner spends effort and risk in clearing the obstacle in exchange for a remaining path along which no failure can cost it the goal.
The residual $4.9\%$ is the probability that a failure occurs during the initial maneuver, while the robot is still inside the obstacle's shadow and no constant-velocity extension of its state reaches $\mathcal{X}_g$.
For those scenarios the mission is lost under any plan. Maximizing \cref{eqn:cost_func} ensures this residual probability is minimized.

\subsection{Hardware Experiments}

Finally, we show our resilient planner on the physical freeflyer platform. 
We first define the input space $\mathcal{U}_{M_i}$ for the platform based on an estimated force per actuator\footnote{The control bounds do not adhere to the subsystems assumption due to air-pressure regulation of the physical platform; however, we note that tightening these bounds would allow the assumption to be satisfied, as it would on many other physical space platforms. }.


\begin{align}
\label{eqn:input_spaces}
    \mathcal{U}_{M_0} &= \left\{ 
        (u_x, u_y, u_{\theta}) \,\middle|\,
        \begin{array}{l}
        u_x \in [-2.125, 2.125] \\
        u_y \in [-2.125, 2.125] \\
        u_{\theta} \in [-0.510, 0.510] \\
        \end{array}
    \right\} \nonumber \\[6pt]
    \mathcal{U}_{M_1} &= \left\{ 
        (u_x, u_y, u_{\theta}) \,\middle|\,
        \begin{array}{l}
        u_x \in [-2.975, 2.975] \\
        u_y \in [0, 0] \\
        u_{\theta} \in [-0.357, 0.357] \\
        \end{array}
    \right\} \nonumber \\[6pt]
    \mathcal{U}_{M_2} &= \left\{ 
        (u_x, u_y, u_{\theta}) \,\middle|\,
        \begin{array}{l}
        u_x \in [0, 0] \\
        u_y \in [-2.975, 2.975] \\
        u_{\theta} \in [-0.357, 0.357] \\
        \end{array}
    \right\}  \nonumber \\[6pt]
    \mathcal{U}_{M_3} &= \left\{ 
        (u_x, u_y, u_{\theta}) \,\middle|\,
        \begin{array}{l}
        u_x \in [0, 0] \\
        u_y \in [0, 0] \\
        u_{\theta} \in [0,0] \\
        \end{array}
    \right\}
\end{align}

The resulting sequence of value functions can be used as a controller according to~\cref{eq:HJB_opt_control} on an accurate model, which is often the case in weightless environments. 
However, due to imperfections in our experimental setup (specifically undulations in the flat epoxy floor), we opt for a more robust control strategy where the value function is used as a reference trajectory generator for a direct-allocation Model Predictive Controller (MPC) as in~\cite{roque2025towards}
This maintains our long-horizon liveness and safety guarantees as the probabilistic guarantees are embedded in the sequential value functions obtained from the planner.

We experimentally verify our algorithm and its reference trajectory on two reach-avoid scenarios.
In both, the failure is injected at $T=2$, i.e. after the initial maneuver that clears the obstacle's shadow; for the plan of \cref{fig:mid_traj} this lies inside the interval over which the success probability stays at its full $0.951$.
These runs are single realizations, intended to show that the planned behavior is executable on hardware.
Injecting the same failure earlier, while the robot is still in the shadow, would leave it drifting along a line that misses $\mathcal{X}_g$ --- the complementary $4.9\%$ of \cref{fig:mid_traj}.

\begin{figure*}[t]
    \centering
    \includegraphics[width=0.96\textwidth]{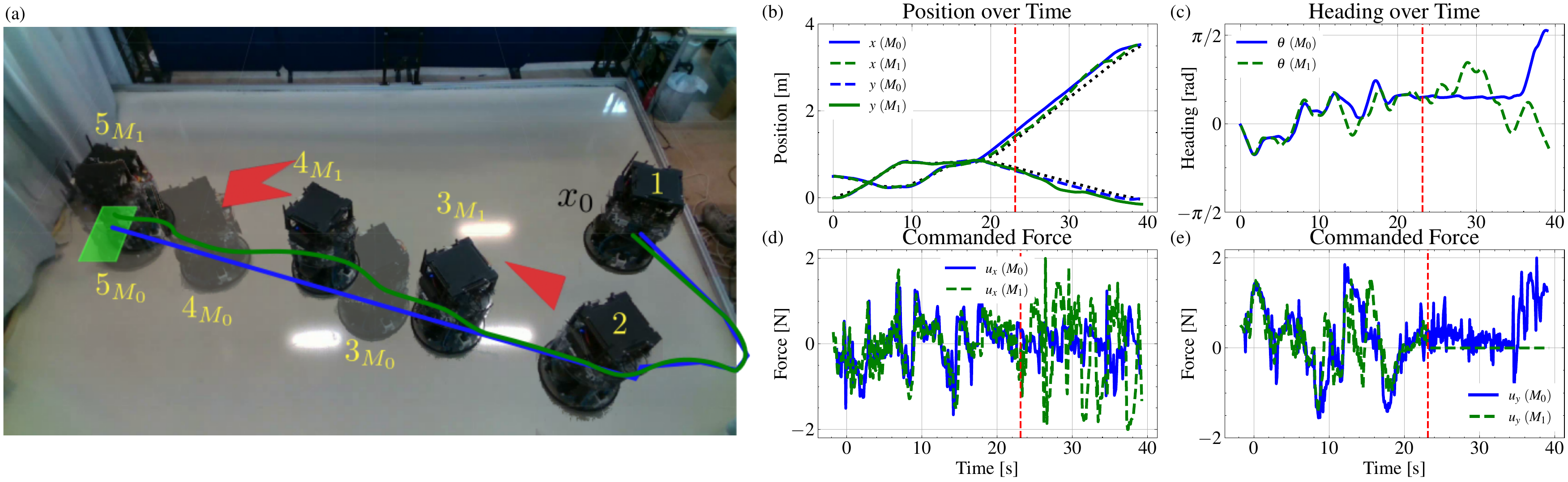}
    \caption{Experimental result of the nominal plan (without failures) and an injected failure at $T=2$ (also denoted by a red vertical line). (a) shows the planned trajectory and the executed trajectory for the nominal model $M_0$ and with a failure model $M_1$. (b) The position over time, showing the increased tracking error after the failure occurs but the recovery after the initial upset. (c) The heading over time where $M_1$ needs to continuously rotate to be able to counteract ground disturbances. (d) The commanded force in the $x$ direction (where no failure occurs). (e) The commanded force in the $y$ direction (where the failure for model $M_1$ occurs).}
    \label{fig:exp_hard_m_1}
\end{figure*}

\begin{figure*}[t]
    \centering
    \includegraphics[width=0.96\textwidth]{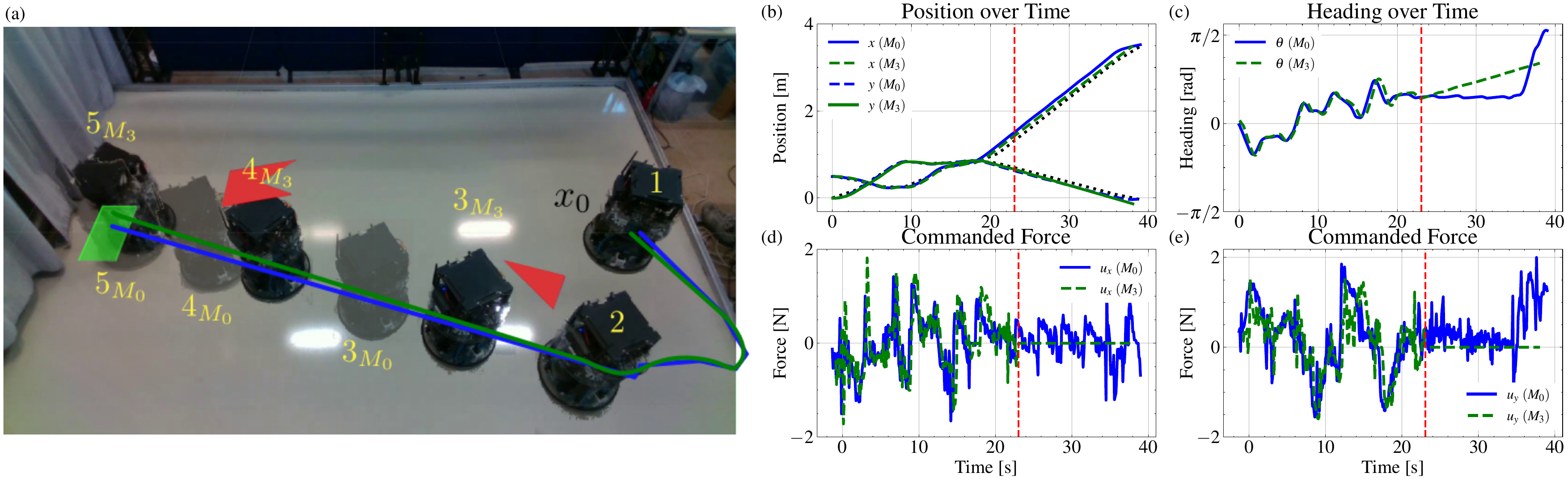}
    \caption{Experimental result of the nominal plan (without failures) and an injected failure at $T=2$ (also denoted by a red vertical line). (a) shows the planned trajectory and the executed trajectory for the nominal model $M_0$ and with a failure model $M_3$ resulting in a complete loss of actuation. (b) The position over time. In contrast to~\cref{fig:exp_hard_m_1} the tracking remains accurate as we lack of actuation is simulated by a velocity tracking MPC initialized at $T=2$. (c) The heading over time where $M_3$ rotates at a constant velocity in weightlessness. (d) The commanded force in the $x$ direction (where the failure occurs). (e) The commanded force in the $y$ direction (where the failure also occurs).}
    \label{fig:exp_hard_m_3}
\end{figure*}

First, \cref{fig:experiment_mid} shows how the reference trajectory from \cref{fig:mid_traj} is executed on the platform under complete actuator failure $M_3$ at $T=2$.
The second scenario is shown in~\cref{fig:exp_hard_m_1} and~\cref{fig:exp_hard_m_3}. 
\cref{fig:exp_hard_m_1} shows the executed trajectory without failures ($M_0$) and with failure mode $M_1$ at $T=2$. 
The failure makes the system non-holonomic, veering the robot off-course (predominantly due to the undulations in the epoxy floor). 
The feedforward roll-out of the value function generates a dynamically feasible reference trajectory, bringing the robot back to its planned trajectory. 
\cref{fig:exp_hard_m_3} shows the behavior under actuator failure model $M_3$ at $T=2$. 

The results highlight the capability of the system to reach its target under partial and complete actuator failures in weightless environments and, importantly, provides a quantitative measure of the system's robustness against such failures.

Note the commanded force plots in~\cref{fig:exp_hard_m_1} and~\cref{fig:exp_hard_m_3} denoting the reduced actuation (i.e. zero actuation on failed axis) for the different failure models.

\section{CONCLUSIONS}



We presented a resilient motion planning framework for spacecraft under actuator failures, combining Markov failure models with backward reachability inside a sampling-based planner, together with the reference policies used to track its plans in hardware.
The resulting motion plans exhibit behaviors that reflect the distinctive dynamics of weightless environments, explicitly trading nominal performance for robustness to actuator loss.

As no plan survives every failure realization, our planner identifies its shortcomings and maximizes the overall probability of success, returning a quantitative measure of the residual mission risk.
The system is represented with all its failure modes with pre-computed reachable sets, making our planner adaptable to different systems and failure scenarios.

Future work will focus on a more thorough analysis of symmetry to enable additional failure models, extensions to higher-dimensional systems, more realistic actuator failure models, the inclusion of orbital dynamics, and a statistical evaluation over a larger set of plans and failure timings.

\bibliographystyle{IEEEtran}
\bibliography{references}

\end{document}